\documentclass[lettersize,journal]{IEEEtran}
\usepackage{amsmath,amsfonts, bm}
\usepackage{amssymb}
\usepackage{algorithm}
\usepackage{array}
\usepackage[caption=false,font=normalsize,labelfont=sf,textfont=sf]{subfig}
\usepackage{textcomp}
\usepackage{stfloats}
\usepackage{url}
\usepackage{verbatim}
\usepackage{graphicx}
\usepackage{cite}
\usepackage{booktabs}
\usepackage{multirow}
\usepackage{amsmath,mathtools}
\usepackage{algpseudocode}
\usepackage{threeparttable}
\usepackage[hidelinks]{hyperref}
\usepackage{balance}
\usepackage{xcolor}
\usepackage{dsfont}

\DeclareRobustCommand{\IEEEauthorrefmark}[1]{\smash{\textsuperscript{\footnotesize #1}}}

\begin{document}

\title{Uncertainty-aware Semi-supervised Ensemble Teacher Framework for Multilingual Depression Detection}

\author{\IEEEauthorblockN{Mohammad Zia Ur Rehman\IEEEauthorrefmark{1}, Velpuru Navya\IEEEauthorrefmark{1}, Sanskar\IEEEauthorrefmark{1}, Shuja Uddin Qureshi\IEEEauthorrefmark{2}, Nagendra Kumar\IEEEauthorrefmark{1}\IEEEauthorrefmark{*}}
   \\ 
    \IEEEauthorblockA{\IEEEauthorrefmark{1}Indian Institute of Technology Indore, India, \IEEEauthorrefmark{2}Shri Govindram Seksaria Institute of Technology and Science, Indore, India
    }
\thanks{\IEEEauthorrefmark{*}Corresponding Author} 
\thanks{All the authors contributed equally to this work.}
% <-this % stops a space
\thanks{Email : phd2101201005@iiti.ac.in (Mohammad Zia Ur Rehman), cse210001075@iiti.ac.in (Velpuru Navya), cse210001063@iiti.ac.in (Sanskar), shujaqureshi2601@gmail.com (Shuja Uddin Qureshi),nagendra@iiti.ac.in (Nagendra Kumar)}% <-this % stops a space
}

\markboth{}%
{}

% % column for its text to clear the IEEEpubid mark.

\maketitle
\thispagestyle{empty}
\begin{abstract}
Detecting depression from social media text is still a challenging task. This is due to different language styles, informal expression, and the lack of annotated data in many languages. To tackle these issues, we propose, Semi-SMDNet, a strong \textbf{\large Semi}-\textbf{\large S}upervised \textbf{\large M}ultilingual \textbf{\large D}epression detection \textbf{\large Net}work. It combines teacher-student pseudo-labelling, ensemble learning, and augmentation of data. Our framework uses a group of teacher models. Their predictions come together through soft voting. An uncertainty-based threshold filters out low-confidence pseudo-labels to reduce noise and improve learning stability. We also use a confidence-weighted training method that focuses on reliable pseudo-labelled samples. This greatly boosts robustness across languages. Tests on Arabic, Bangla, English, and Spanish datasets show that our approach consistently beats strong baselines. It significantly reduces the performance gap between settings that have plenty of resources and those that do not. Detailed experiments and studies confirm that our framework is effective and can be used in various situations. This shows that it is suitable for scalable, cross-language mental health monitoring where labelled resources are limited.
\end{abstract}

 
\section{Introduction}
\label{sec:introduction}

Social media platforms are a big part of daily life. They give users ways to express themselves, communicate, and connect. However, these platforms also show the mental states of users, including signs of psychological distress like depression. Recognizing depressive signs in social media posts can help prompt timely interventions. This can reduce the risk of serious outcomes like self-harm~\cite{anshul2023multimodal}. Despite the significant societal implications, accurate depression detection remains a challenging task due to the multilingual and informal nature of social media content.

\noindent
Traditional approaches for depression detection rely heavily on supervised machine learning methods that require large amounts of labeled data~\cite{hasib2023depression}. However, for low-resource languages, the lack of annotated datasets hinders the development of robust models. Multilingual pre-trained models such as mBERT and XLM-R provide a foundation for natural language processing (NLP) tasks, but their performance is often suboptimal for dialect-rich and code-mixed texts common on social media~\cite{guo2023prompt, ghosh2023attention}. The linguistic diversity, combined with inconsistencies in Romanized scripts and spelling variations, further complicates the task~\cite{figueredo2022early}.

\noindent
The global rise in internet and smartphone use has led to a big increase in social media content. For instance, in India, more than 759 million internet users actively engage with platforms like Twitter, Facebook, and WhatsApp. They often communicate in several languages or mixed text~\cite{IAMAI2022Internet, DataReportal2024India}. Code-mixed content occurs when users blend languages. This creates unique challenges because of different grammatical rules and informal language patterns~\cite{rehman2023user}. These differences complicate depression detection models, which makes traditional methods inadequate.

Recently, semi-supervised learning methods have been recognised as effective solutions to the issue of having language datasets with few resources. Pseudo-label revising methods have been introduced, and their performance has been very impressive, particularly in situations that are linguistically diverse and low-resource~\cite{almutairi2025flick}. Despite these developments, depression detection in multiple languages still remains a complicated problem due to the existence of several significant challenges. First of all, low-resource languages that lack enough annotated data make it difficult to apply supervised training effectively. Secondly, the informal and dialect-rich nature of social media can cause linguistic variation, which in turn makes it hard for standard models to generalise. Thirdly, semi-supervised techniques that are used optimistically may produce noisy pseudo-labels that can confuse the learning process. Furthermore, there are issues regarding scalability if these methods are used on devices with limited memory and processing power, and serious privacy concerns arise when dealing with sensitive mental health data. Collectively, these problems call for robust and versatile modelling strategies. 

Our approach tackles these challenges with improved pseudo-labelling, uncertainty-based filtering, and data augmentation. This enables dependable and inclusive multilingual depression detection. We propose a semi-supervised learning framework for multilingual depression detection on social media tweets. Our approach combines state-of-the-art transformer-based models for text embeddings with student-teacher models for robust pseudo-labelling. We enhance the framework with ensemble learning techniques and advanced data augmentation strategies to handle linguistic diversity effectively and to enrich the predictive capability of our system.

\subsection{Key Contributions}

The following points highlight the key contributions of our research:

\begin{enumerate}
\item The paper introduces a semi-supervised framework for multilingual depression detection that uses teacher-student pseudo-labeling and data augmentation. It has been tested on Arabic, Bangla, English, and Spanish datasets. This method consistently beats the baseline results and reduces the performance difference between resource-rich and resource-poor settings.

 \item An ensemble of XLM-RoBERTa teacher models generates pseudo-labels through soft voting. An uncertainty threshold filters out low-confidence predictions. This reduces label noise and improves the learning process's stability. It also enhances label quality and learning stability.
 
\item A confidence-based weighting system emphasizes trustworthy pseudo-labels during training. An ablation analysis shows that this method significantly improves robustness in multilingual datasets.

\item Comprehensive experiments prove the framework's strong performance and generalizability. Cross-lingual results confirm that refining pseudo-labels, using ensemble learning, and applying uncertainty weighting together lead to solid improvements, especially in low-resource situations.

\end{enumerate}

\section{Related Work}
\label{sec:related_works}

Early efforts to detect depression from social media mainly looked at English-language content. Researchers used NLP and machine learning techniques to spot signs of depression in user posts~\cite{Choudhury2013PredictingDV,eichstaedt2018facebook}. However, cultural and language differences significantly affect how people express depression online.

Recent studies have shifted to multilingual and cross-lingual models. Khalil et al.~\cite{khalil2024federated} used federated learning for multilingual depression detection. They pointed out the privacy advantages and the difficulties of dealing with language imbalance. Bucur et al.~\cite{bucur2025survey} provided a wide-ranging survey of multilingual mental health detection. They focused on dataset diversity, annotation practices, and the need for clear methods.

Language-specific research includes RNN-based and transformer-based models for Bengali~\cite{ahmed2024depression,kabir2022detection}, lexical and deep models for Arabic~\cite{alghamdi2020predicting,helmy2024depression}, and annotated benchmarks for Spanish through shared tasks like MentalRiskES~\cite{romero2024mentalriskes} and user timeline analysis~\cite{leis2019detecting}. Explainable AI has also gained attention, especially for Bengali and Spanish datasets~\cite{ghosh2023attention,bao2024explainable}.

Transformers such as Multilingual BERT and XLM-Roberta, along with language-specific versions like AraBERT and Bangla-BERT, are now commonly used for both monolingual and cross-lingual depression detection~\cite{kabir2022detection,bucur2025survey}. However, challenges remain with resource scarcity and linguistic diversity.

More recently, large language models (LLMs) like GPT-4 and specialized mental-health-focused versions have shown promise in multilingual settings~\cite{xu2025evaluation,xu2024mental}. LLMs exhibit strong diagnostic accuracy and generalization, especially when improved with techniques like chain-of-thought prompting~\cite{teng2025enhancing} or multimodal inputs~\cite{sadeghi2024harnessing}. This development opens the door for more scalable and understandable detection systems.

\section{Methodology}
\label{sec:methodology}
\begin{figure*} % Use [H] to force placement
    \centering
    \includegraphics[width=1.0\linewidth]{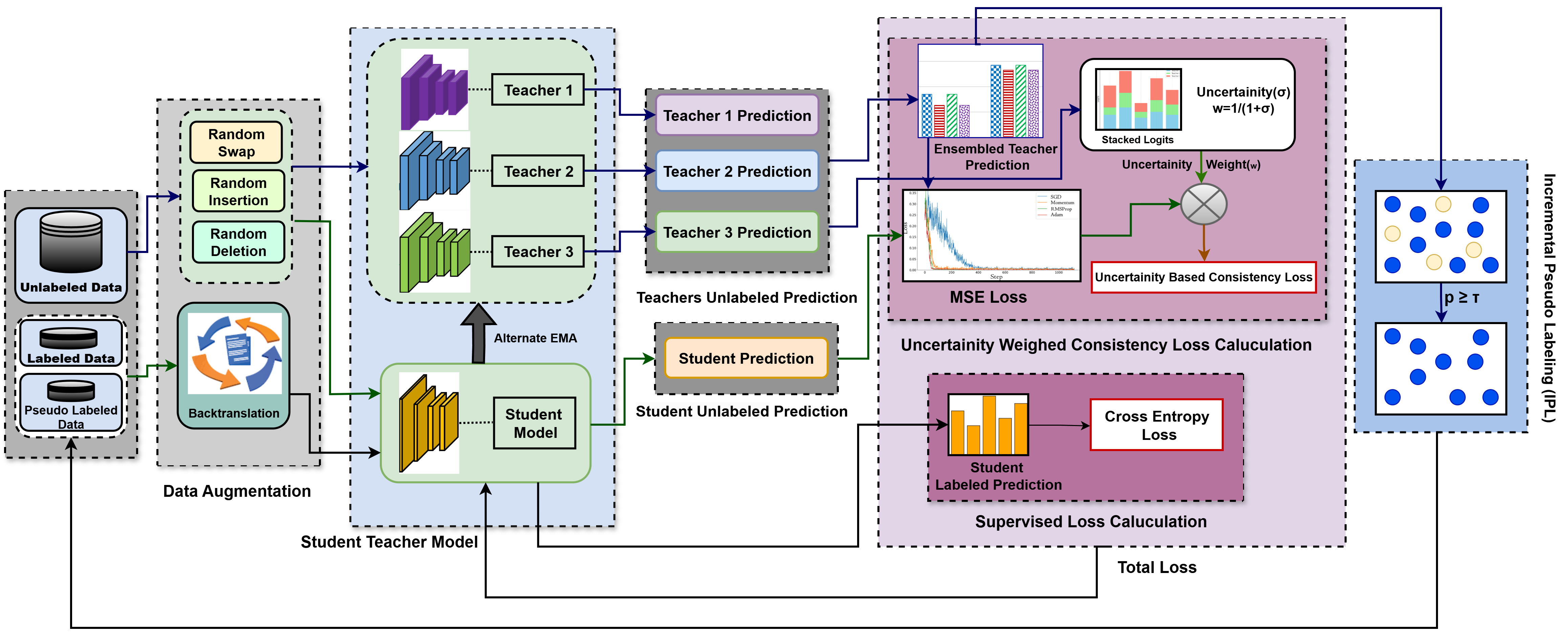}
    \caption{System Architecture}
    \label{fig:architecture}
\end{figure*}

This section provides a description of our method, Semi-SMDNet, for multilingual depression detection through semi-supervised learning. The system design, as shown in \autoref{fig:architecture}, has five key modules: (a) Multilingual Dataset Preparation, (b) Data augmentation, (c) Student-Teacher Framework, (d) Loss Functions and Training Approach, and (e) Incremental Pseudo-Labelling (IPL). The method starts with the creation of a diverse dataset in several languages, which is then used in a semi-supervised learning setting with a teacher-student framework to improve classification performance. 
The training objective is to include the optimisation of supervised and consistency loss functions, with special emphasis on uncertainty-based measures to fine-tune model predictions. Besides that, a novel IPL strategy is presented to gradually update the labelled and unlabeled data pools. In the end, data augmentation is used to enhance the model's capability. The details of each of these components are provided ‌below. 

\subsection {\textit{Multilingual Dataset Preparation}}
We have put together an extensive dataset for this research which includes four languages: Bangla, Arabic, English, and Spanish. These languages are chosen as representatives of distinct language families. Each of them has its own syntactic rules, orthographic systems, and even its own culture. This linguistic diversity sets a wide array of problems that the semi-supervised approach we have developed is capable of solving, for instance, dealing with differences in script, vocabulary, and grammatical structures. 

The dataset is a mixture of both formal and informal texts from various sources, thus representing both formal and informal language usages.  Data from social media, online forums, and community interactions contains informal language, regional dialects, and colloquial expressions which normally are not accounted for in traditional NLP datasets. This variety not only makes the models more powerful but also makes them more flexible to be used in the real world, where informal and code-mixed languages are common. 

\subsubsection{\textit{Dataset Overview}} 
 For this study, we gathered data in four distinct languages: Bangla, Arabic, English, and Spanish. These languages were chosen due to their varied grammatical rules, syntactic complexities, and rich cultural contexts, posing unique challenges for natural language processing models. Details of datasets are given in Tables \ref{tab:arabic_dataset}, \ref{tab:bangla_dataset}, \ref{tab:english_dataset}, and 
\ref{tab:spanish_dataset}.
\begin{enumerate}
    \item \textbf{Bangla, Arabic, and English Datasets} \\
    We leveraged pre-existing collections from prior works for the Bangla~\cite{uddin2019depression}, Arabic~\cite{helmy2024depression}, and English datasets~\cite{helmy2024depression}. These datasets were deliberately selected to cover different linguistic changes in languages, such as the use of informal languages, dialects, and regional slangs. The Bangla dataset, as per the references in the research, is a mixture of the formal and colloquial expressions, which are generally the social interactions. The Arabic dataset consists of the samples of the Modern Standard Arabic (MSA) and the dialects of the regions like Egyptian and Levantine Arabic, thus, the language diversity and the complexity being reflected as per the description in the researched articles. The English dataset also, as derived from the research sources, comprises the texts of different informal environments, where the slang, abbreviations, and the non-standard grammatical structures are used. These datasets are extensively confirmed in scholarly research, thus, they are dependable resources for the development of strong language models. 

    \item \textbf{Spanish Dataset} \\
    The Spanish dataset~\cite{leis2019detecting} was a handpicked selection of the content of social media platforms and online community forum discussions. The gathering of this data was aimed at getting the language variants of Spanish from both Europe and Latin America so that the dataset would have the linguistic features of the two regions. The Spanish dataset, therefore, is a reflection of the language used in the social networks which is an informal and diverse one, as it has been derived from real-world, user-generated content. The manual curation of our dataset was a way of making it compatible with the requirements of our semi-supervised learning framework, thus, it could cope with the intricacies of informal digital communication. 

    \item \textbf{Multilingual Diversity and Balance} \\
    The entire data set was designed to keep a balanced representation of each language so that our models would be able to generalize in different linguistic contexts. Such a multilingual strategy is a step towards models which can face various language challenges and at the same time, it provides a way to measure the effectiveness of semi-supervised learning methods in using unlabeled data in different language environments. We have, thus, gathered a diverse, heterogeneous collection that is fit for the scalability and robustness of our learning models to be tested by merging datasets from both academic sources and the manually curated social media content. 
\end{enumerate}

\begin{table*}[h!]
\centering
\begin{tabular}{cc}
% ---------- Row 1 ----------
\begin{minipage}{0.45\linewidth}
\centering
\caption{Arabic dataset details.}
\begin{tabular}{|l|l|}
\hline
\textbf{Class} & \textbf{Number of Samples} \\ \hline
Non\_depressed & 5,000 Tweets \\ \hline
Depressed & 5,000 Tweets \\ \hline
\end{tabular}
\label{tab:arabic_dataset}
\end{minipage}
&
\begin{minipage}{0.45\linewidth}
\centering
\caption{Bangla dataset details.}
\begin{tabular}{|l|l|}
\hline
\textbf{Class} & \textbf{Number of Samples} \\ \hline
Non\_depressed & 2,930 Tweets \\ \hline
Depressed & 984 Tweets \\ \hline
\end{tabular}
\label{tab:bangla_dataset}
\end{minipage}
\\[1.2em] % Space between rows
% ---------- Row 2 ----------
\begin{minipage}{0.45\linewidth}
\centering
\caption{English dataset details.}
\begin{tabular}{|l|l|}
\hline
\textbf{Class} & \textbf{Number of Samples} \\ \hline
Non\_depressed & 38,219 Tweets \\ \hline
Depressed & 21,953 Tweets \\ \hline
\end{tabular}
\label{tab:english_dataset}
\end{minipage}
&
\begin{minipage}{0.45\linewidth}
\centering
\caption{Spanish dataset details.}
\begin{tabular}{|l|l|}
\hline
\textbf{Class} & \textbf{Number of Samples} \\ \hline
Non\_depressed & 1,186 Tweets \\ \hline
Depressed & 1,000 Tweets \\ \hline
\end{tabular}
\label{tab:spanish_dataset}
\end{minipage}
\end{tabular}
\end{table*}

\subsubsection{\textit{Data Preprocessing Steps
}}
 Preprocessing of a multilingual dataset is a very crucial step that ensures data consistency and high-quality data, especially when the sources are too informal. The raw text data of each language have gone through an extensive normalization process. At the very beginning, all the text is converted into lowercase so that uniformity is maintained. Besides that, unnecessary characters such as special symbols, URLs, and redundant whitespaces are removed from the text. This normalization is a necessity as it helps to get rid of the noise in the dataset, thus it becomes easier for models to learn the patterns that are of real value. Each of the languages has its own peculiarities and characteristics that is why language-specific preprocessing methods have been used. For example, the removal of stopwords was performed by using the predefined lists of NLTK libraries, which were applied for Bangla, Arabic, English, and Spanish so as to ensure the account of commonly used functional words that do not add anything to the semantic meaning. Besides that, emojis, and emoticons that are very common in social media text, were transformed into the identical textual descriptions. By doing this, the sentiment conveyed by these visual symbols is not lost but rather the dataset is enriched with more expressive content.

One of the major issues in the preprocessing stage was how to deal with the differences in dialects, especially for languages such as Arabic and Spanish whose vocabularies vary greatly depending on the region. To solve that, we used lexicons to convert colloquial words into their formal equivalents, thus lessening the variability of the dataset, however, at the same time, retaining its semantic richness. The whole preprocessing pipeline has been sufficiently done in such a way that the data used for the semi-supervised learning model is not only clean and consistent but also representative of the real use of language in the world. 

\subsubsection{\textit{Data Splits (Labeled, Unlabeled, Test)
}} 
The success of a semi-supervised learning pipeline depends largely on how well the dataset is divided into labeled, unlabeled, and test portions. In this research, we used a clever approach in breaking down the data so that it could be used efficiently across different languages and at the same time, the linguistic diversity would be equally represented.

The labeled data is a subset of the original multilingual corpus that has been thoroughly annotated. Based on a stratified sampling method, we chose this subset so that each language would be represented in proportion to its frequency in the raw dataset. It was a very important step in removing the risk of any language dominating the labeled data, hence, biasing the model during the training phase. In general, the annotations were made by labeling the data points with the respective class labels through a mixture of human annotators and automated tools to maintain the accuracy at a high level. The human annotators were either native speakers or proficient in the respective languages, hence, the chances of misinterpretations, especially for the context-sensitive slang and idiomatic expressions, were minimal.

The unlabeled data is basically the bulk of our training data and is a mirror of the real-world situation where labeled data is limited. This dataset is made up of raw text data without explicit class labels and, therefore, contains a lot of contextual information that the semi-supervised model can utilize to enhance its performance. The unlabeled fraction was deliberately left large, making up about 0.6 of the total dataset, so that the model’s potential in learning from unannotated data could be fully realized. The variety in this dataset was secured by having examples from various domains such as news articles, social media posts, and online discussions, which, in turn, guarantees a large number of linguistic structures and contexts.

We assembled the test dataset as an unbiased standard for measuring the performance of the model. The set is completely different from both the labeled and unlabeled training datasets, which means that the evaluation is not influenced by any data that the model has previously seen. To ensure a realistic evaluation, the test set contains samples from the same sources as the training data, and these cover different language registers and domains. Moreover, the test set was also balanced in terms of language distribution so that the evaluation metric would be equally fair for all four languages.

The ultimate data split was close to 0.2 for labeled data, 0.6 for unlabeled data, and 0.2 for testing, which means that the model gets to have a sufficient amount of unlabeled data to learn from as well as a test set that is large enough to evaluate its generalization capabilities.

\subsection {\textit{Semi-Supervised Learning Framework
}} \label{section:Semi-Supervised Learning Framework}
The semi-supervised learning framework this work used was mainly a student-teacher model that by design is able to adapt to usage of both labeled as well as unlabeled data. The main concept is that the teacher model which is initially trained on a small set of labeled data is applied on the unlabeled data for generating the pseudo-labels. The newly pseudo-labeled data are then used for training the student model which hence tries to exceed the teacher's performance by having access to a more extensive dataset for learning. Such an approach is essentially advantageous in multilingual contexts where it happens that there are few labeled data for some languages and thus, the model can still make effective use of the unlabeled data to improve its generalization capacity.
\subsubsection{\textit{Overview of the Student-Teacher Model}} 
Our semi-supervised learning framework mainly revolves around the student-teacher paradigm, which is a hybrid approach that uses both labeled and unlabeled data to improve the model's performance. We apply the same architecture, XLM-RoBERTa (Cross-lingual Language Model RoBERTa), for the teacher and student models, in this case. The main reason for the application of XLM-RoBERTa is its efficient capability in handling multilingual data, owing to its ability to detect very detailed linguistic patterns for different languages.

The labeled subset of our multilingual dataset that consists of samples in Bangla, Arabic, English, and Spanish is used in the very beginning to pre-train the teacher model. This pre-training step deals with adjusting the pre-trained XLM-RoBERTa model to represent the languages in the corpus at a higher level. The teacher model is therefore employed to annotate the unlabeled dataset after this stage. These pseudo-labels are basically the predicted classes along with the confidence scores, which indicate the teacher model's confidence in the predictions.

Only those pseudo-labels that have high confidence scores are used to train the student model so as to ensure their quality. The process of selection here is instrumental in not only minimizing the noise in the learning input but also in maximizing it. The practice of omitting low confidence predictions allows the model to give full play to the most trustworthy data, which is very important for its performance leverage in a semi-supervised setting.

Distinguished into two parts, the student model is also based on XML-RoBERTa. Firstly, on the same labeled data that were used for the teacher model, the student model undergoes training and thus gets initialized with the strong baseline. Next, the student model is trained on the high-confidence pseudo-labeled data generated by a teacher model. A two-step learning method helps the student machine harness more extensive and diverse data, that is, it skillfully realizes the power of the semi-supervised learning paradigm. Besides, the student model training comprises conforming regularity which guarantees that the model's output stays unaltered by different kinds of modifications of the input data such as data augmentations and noise injections.

This student-teacher scheme uses an iterative strategy where the student model is in a position to become better than the original teacher model through additional pseudo-labeling.

The student model, after being trained, can therefore be considered as a new teacher in the following iterations which is a way of progressively enhancing the pseudo-labeling process. The iterative refinement, in particular, has a potent impact on not only increasing the robustness of the model but also its generalization abilities across the multilingual datasets.

\subsubsection{\textit{Ensemble of Teachers (3T)}}
We got the pseudo-labels to be more robust and trustworthy by setting up a teacher model ensemble. Our 3T (Three Teacher) framework comprises three instances of the XLM-RoBERTa models, each with a different seed for initialization and slightly different hyperparameters. The idea was to add some variation between the teachers so that the bias that comes from being too much influenced by a single model's idiosyncrasies is reduced and so the student model receives a richer and more reliable set of pseudo-labels.

The individual predictions of the three teachers are put together using a soft voting mechanism. This procedure assumes that each of the three teacher models not only predicts a class, but also outputs a confidence score, and these scores are averaged to yield the final pseudo-label. We also employ an uncertainty threshold that discards the pseudo-labels for which the confidence scores from the ensemble are lower than a predefined level. This filtering operation ensures that the student model only receives some high-quality pseudo-labeled data, thus the training process is further enlightened.

There are numerous benefits that come into play when an ensemble of XLM-RoBERTa models is used, especially in a multilingual setting where language diversity is the source of the most significant challenges. The ensemble approach stabilizes the training procedure, lessens the prediction variance, and increases generalization and robustness. In this way, the combined strengths of several models enable us to efficiently deal with the complexities of multilingual datasets.

\begin{table*}[h!]
\centering
\caption{Tuning range and configurations for the CustomTeacherModel and Student Model.}
\small % Reduce the size of the table
\begin{tabular}{|l|c|l|}
\hline
\textbf{Layer/Component}        & \textbf{No. of Layers Tried} & \textbf{Tuning Range}                   \\ 
\hline
\text{XLM-Roberta Encoder}    & 12 (fixed)                  & Pretrained embeddings, 768 dimensions        \\ 
\hline
\text{Dropout}                & 1 per model                 & 0.1–0.3 (Teachers), 0.1 (Student)            \\ 
\hline
\text{Dense (Softmax Activation)} & 1 (fixed)                  & 2 neurons (binary classification)            \\ \hline
\text{Teacher Models}         & 3                           & Seeds: [42, 43, 44]; Dropout: [0.1, 0.2, 0.3] \\ \hline
\text{Student Model}          & 1                           & Pretrained with XLM-RoBerta base             \\ 
\hline
\text{Regularization}         & Implicit                    & Handled by pretrained XLM-RoBerta layers     \\
\hline
\end{tabular}
\label{tab:model_tuning}
\end{table*}

This ensemble-based student-teacher model forms the foundation of our semi-supervised learning framework, enabling us to harness the full potential of both labeled and unlabeled data in multiple languages.

\subsubsection{Exponential Moving Average (EMA)}
In our semi-supervised learning framework we use an Exponential Moving Average (EMA) technique to make the student-teacher model architecture more stable and to increase its performance. EMA is a favorite method in semi-supervised learning, especially in teacher-student frameworks, to refine the pseudo-labels that the teacher model generates. As a result, the model yields more consistent and accurate predictions throughout the training process since it averages the parameters of the model over time.

The main point of EMA is to have a different set of model parameters for the teacher model that are updated as an exponentially weighted moving average of the student model's parameters. Simply put, the parameters of the teacher model are a more stable version of the parameters of the student model (which rapidly change during training) that have been smoothed over several training iterations, and hence, the teacher model is not directly using the parameters of the student model. Making the pseudo-labels more stable, thus, ensures that they can be used efficiently in the semi-supervised learning process.

Let $\theta_t$ represent the parameters of the teacher model at training step $t$, and $\theta_s$ represent the parameters of the student model. The update rule for the teacher model’s parameters using EMA can be defined as:
\begin{equation}
\theta_{t} = \alpha \theta_{t} + (1 - \alpha)\theta_{s}
\end{equation}
where, $\alpha \in [0, 1]$ is the decay rate, a hyperparameter that controls the extent of smoothing. A higher $\alpha$ results in more stable but slower updates, while a lower $\alpha$ allows the teacher model to adapt more quickly to changes in the student model. $\theta_t$ is initialized to be equal to $\theta_s$ at the start of training.

The choice of $\alpha$ is critical for the balance between stability and adaptability. Empirically, values like $\alpha=0.99$ or $\alpha=0.999$ are found to work well, as they provide sufficient smoothing without overly lagging behind the student model.

When training is going on, the student model's parameters are changed in the usual way by backpropagation and gradient descent. After every change, the teacher model's parameters are changed by the EMA formula given above. This method is fast in terms of calculations because it only needs a simple weighted sum update which has a very small overhead if compared to the gradient descent step.

Also, the EMA teacher model is there only to help the training process by creating the pseudo-labels. At inference, we take the student model that has been trained with the help of the teacher and thus can be considered as reliable.

\subsection{Loss Functions and Training Approach}
Our model's training system relies on a mix of supervised and unsupervised loss functions, which are aimed at using both labeled and unlabeled data to the maximum extent. The idea is to improve the model as much as possible with the available labeled data, and at the same time, to use the huge amount of unlabeled data to make the model more general. Our method, which is an implementation of cross-entropy loss for the labeled examples and a consistency loss for the unlabeled ones, encourages correctness as well as strength of the model. We explain there the parts of the loss functions employed.
\subsubsection{Supervised Loss Calculation}
At the core of our training methodology, we initially consider a conventional supervised loss that makes use of the labeled subset of the dataset. To measure the difference between the predicted class probabilities and the ground truth labels, the cross-entropy loss function is utilized. This method works especially well for multi-class classification issues since it penalizes incorrect predictions and gives rewards to correct ones, thus, it helps the model to achieve a higher accuracy level. The cross-entropy loss for each instance is calculated as:
\begin{equation}
L_{\text{supervised}} = -\sum_{c=1}^{C} y_{c} \log(\hat{y}_{c})
\end{equation}
where $C$ represents the number of classes, $y_{c}$ is the true label (one-hot encoded), and $\hat{y}_{c}$ is the predicted probability for class $c$. By minimizing this loss over the labeled dataset, we establish a strong baseline for the model’s performance on labeled examples.

\subsubsection{Consistency Loss Calculation}
In order to make maximum use of the unlabeled dataset, a consistency loss is presented, that was the main factor in enabling prediction stability to be raised considerably by the model. This loss is the one that makes the student model keep its predictions in line with the teacher versions ensemble predictions on the unlabeled data. The consistency loss is formed by the mean squared error (MSE) between the logits of the student model and the averaged logits of the three teacher models.

The consistency loss is defined as:

\begin{equation}
L_{\text{mse}} = \left\| f_{\text{student}}(x) - \frac{1}{3} \sum_{i=1}^{3} f_{\text{teacher}_{i}}(x) \right\|^2
\end{equation}

where, \(x\) represents the weakly augmented unlabeled input (e.g., via random swapping, deletion, or insertion of tokens),  \(f_{\text{student}}(x)\) is the output of the student model for input \(x\), and \(f_{\text{teacher}_{i}}(x)\) denotes the logits output from the \(i\)-th teacher model. The term \(\frac{1}{3} \sum_{i=1}^{3} f_{\text{teacher}_{i}}(x)\) represents the averaged logits of the three teacher models.

This approach enforces the student model to learn from the generalizations of the ensembled teacher models. Weak augmentations applied to the unlabeled dataset help maintain semantic consistency while exposing the model to input variability. By minimizing the MSE between the student and ensembled teacher logits, the consistency loss improves the robustness and generalization capabilities of the student model.

\subsubsection{Uncertainty Measurement and Total Loss Calculation}
One of the notable characteristics of our semi-supervised learning framework is the incorporation of an uncertainty-based weighting method to make the training process more efficient, especially when working with the data annotated by a model. Since the correctness of the pseudo-labels is always different, it is very important to set their impact according to the model's certainty. We use a committee of teacher models to measure this uncertainty and thus change the usage of each instance with a pseudo-label.
\paragraph{Mathematical Formulation of Uncertainty}
Let $\mathbf{z}_1, \mathbf{z}_2, \mathbf{z}_3$ represent the logits (predicted scores before applying softmax) produced by the three teacher models for a given input. We quantify the uncertainty using the variance of these logits, capturing the degree of agreement among the teacher models:
\begin{equation}
\text{Variance}(\mathbf{z}) = \frac{1}{3} \sum_{i=1}^{3} (\mathbf{z}_i - \bar{\mathbf{z}})^2, \quad \text{where } \bar{\mathbf{z}} = \frac{1}{3} \sum_{i=1}^{3} \mathbf{z}_i
\end{equation}
The mean-variance across all class logits provides the overall uncertainty measure:
\begin{equation}
\text{Uncertainty} = \frac{1}{C} \sum_{c=1}^{C} \text{Variance}(\mathbf{z}_c)
\end{equation}
where $C$ is the number of classes.

Using this uncertainty measure, we define an uncertainty weight:
\begin{equation}
w_{\text{uncertainty}} = \frac{1}{1 + \text{Uncertainty}}
\end{equation}
This weight is high when the variance is low (indicating high confidence) and decreases as uncertainty increases. This adaptive weighting strategy ensures that high-confidence pseudo-labels have a greater impact on the training process, while low-confidence predictions are down-weighted.

\paragraph{Total Loss Calculation}
The total loss $L_{\text{total}}$ for training the student model is a combination of the supervised loss from the labeled data and the consistency loss from the pseudo-labeled data, adjusted by the uncertainty weight:
\begin{equation}
L_{\text{total}} = \alpha L_{\text{supervised}} + \beta w_{\text{uncertainty}} L_{\text{consistency}}
\end{equation}
where, $L_{\text{supervised}}$ is the cross-entropy loss calculated on the labeled dataset, $L_{\text{consistency}}$ is the consistency regularization loss, encouraging stable predictions across different data augmentations. $\alpha$ and $\beta$ are hyperparameters that control the balance between supervised and unsupervised components of the training.

By incorporating the uncertainty weight into the consistency loss, our model prioritizes learning from high-confidence pseudo-labels, thus improving the effectiveness of the semi-supervised learning framework. This strategy not only enhances the robustness of the model but also ensures better generalisation across the diverse multilingual datasets in our study.

\subsection{Incremental Pseudo-Labelling (IPL)}
The IPL technique is a crucial element of our semi-supervised learning framework, designed to make efficient use of large volumes of unlabeled data by selectively incorporating them into the training process\cite{zou2024incremental}. In contrast to conventional pseudo-labeling methods that designate labels for the full unlabeled dataset at once, IPL brings in a gradual and controlled labeling mechanism. Through the use of a dynamic confidence threshold, the model is able to gradually label only those predictions that it is most confident about, thus reducing the chances of error-propagation of incorrect labels. This approach guarantees that the model is initially working with a very accurate, though small, set of pseudo-labeled samples which is gradually expanded as the model becomes more skilled, thus achieving a balance between accuracy and coverage of the training data.

\subsubsection{Thresholding for Label} The IPL strategy aims at making the best use of unlabeled data by progressively adding to the training set those pseudo-labelled samples that have the highest confidence. The operation is controlled by a confidence threshold device, that is, only those predictions that have a confidence score above a previously set threshold are chosen as pseudo-labels. The reasoning for this is to reduce as much as possible the contaminating noise caused by wrong labels that can deteriorate the model training.

Initially, the teacher model generates class probabilities for each sample in the unlabeled dataset. These probabilities serve as a proxy for the model’s confidence in its predictions. The threshold $\tau$ is set to a high value (e.g., 0.92) in the early stages of training, ensuring that only highly confident predictions are considered. As the training progresses, $\tau$ is gradually lowered to include more samples, thus expanding the labeled dataset incrementally. This adaptive thresholding allows the model to start with a small, highly reliable set of pseudo-labelled examples and progressively include more diverse samples, which enhances its generalisation capabilities. The choice of threshold is critical: too high a threshold results in a limited set of pseudo-labels, thereby underutilizing the unlabeled data, while too low a threshold can introduce noisy labels, leading to performance degradation. Therefore, a dynamic approach is employed where the threshold adapts based on the model's training accuracy on the labeled set. This ensures that the expansion of pseudo-labels aligns with the model’s learning capacity.

\subsubsection{Updating Labeled and Unlabeled Sets}
Under the IPL setting, the division of labeled and unlabeled data changes refer to the ongoing updating process that is continuously carried out in the training process. The samples that meet the confidence threshold from the unlabeled set after each epoch are actually those that are transferred to the labeled set, thus making them the new training examples with pseudo-labels indirectly. The data's dynamic reallocation therefore allows for a maximum inclusion of the available information and supplies the model with a tool by which it can learn orderly from the progressively difficult examples.

In case the labeled set gets so rich in pseudo-labeled data, we are taking a measure of setting a limit on the proportion of pseudo-labeled samples there, thus ensuring a proper balance between human-labeled and pseudo-labeled data. The limit that we introduced here prevents the model from becoming biased towards its own predictions, hence the training process stays strong.

When the sample is transferred to the labeled set, it is sometime re-checked in the next training epochs, besides, the checking is done periodically. If the model's confidence in the pseudo-label drops, the sample may be sent back to the unlabeled set so that labeling can be corrected. The model's ability to switch between the labeled and unlabeled sets, therefore, becomes more flexible, especially when it deals with noise or ambiguous data.

The IPL scheme not only speeds up the model's learning journey, but also solves the issue of label scarcity that is more likely to happen in low-resource language areas. By gradually adjusting the pseudo-labeling process, the model is able to get a better grasp of complicated language structures which in turn results in a more accurate and stronger multilingual model.

\subsection{Data Augmentation Techniques}

Without data augmentation, our semi-supervised learning framework would not be as effective. Data augmentation is, thus, the main factor in the model's robustness and generalization capabilities. Our artificial expansion of the training dataset is intended to reduce overfitting and improve the model's capacity to cope with language variability, which is a natural way of the dataset being multilingual. Essentially, our augmentation plan has the following two main goals: first, to enlarge the data diversity, and, second, to use perturbation-based learning to ensure that the model predictions are consistent.
\subsubsection{Overview of Augmentation Strategy}
Our augmentation pipeline incorporates easy data augmentation such as random swap and insertion of synonyms. We also utilize back translation technique. These strategies are meticulously designed to cater to the unique characteristics of multilingual datasets, ensuring effective learning from diverse linguistic inputs.

\begin{itemize}
    \item \textbf{Random Swap, Insertion, and Deletion:} In order to vary syntax, the variations are introduced by means of random word swapping, insertion, and deletion within sentences. These changes can change the word order or add the new context without changing the meaning to a great extent. These kinds of augmentations are especially effective for the case of informal language structures and regional dialects.
    
    \item \textbf{Backtranslation:} In order to have a strong data augmentation technique, we use backtranslation which is a part of the process where a text is translated to an intermediate language and then back to the original language. The aim here is to keep the semantics the same while providing textual variants that are rephrasings of the original input. The application of such a method is to an extent extremely useful for the language model to have the ability of transfer learning between different languages having various syntactic structures that in turn leads to better performance in languages with fewer data.

\end{itemize}

Using data augmentation has a major positive impact on the performance of a semi-supervised model, especially in the case of low-resource languages that have very little labelled data. The main reason for this is that by creating more variability in the training data, the model augmentation can learn the language structure and the meaning of the words that are the core of the language and therefore have better generalisation ability.

\section{Experimental Evaluations}
This section shows the experimental results to demonstrate the effectiveness of our technique.
The results presented in Table \ref{tab:results} showcase the comparative performance of several models across different languages for depression detection in social media text. This analysis focuses on key models, including GPT-4~\cite{gpt}, Hybrid Explainability (IIE)~\cite{ghosh2023attention}, and LLM Semi-supervised (LLM-SS)~\cite{farruque2024depression}, alongside baseline models such as Multilingual BERT~\cite{devlin-etal-2019-bert} and XLM-RoBERTa~\cite{conneau2020unsupervised}, to provide a comprehensive understanding of their effectiveness in this domain.

\begin{table*}[h!]
\centering
\scriptsize
\caption{Performance comparison across languages using 100\% and 20\% labelled data.}
\begin{tabular}{|l|l|c|c|c|c||c|c|c|c|}
\hline
\textbf{Language} & \textbf{Model} &
\multicolumn{4}{c||}{\textbf{100\% Data}} &
\multicolumn{4}{c|}{\textbf{20\% Data}} \\
\cline{3-10}
& & \textbf{Acc} & \textbf{P} & \textbf{R} & \textbf{F1} &
       \textbf{Acc} & \textbf{P} & \textbf{R} & \textbf{F1} \\ \hline

% ================= Arabic =================
\multirow{6}{*}{\textbf{Arabic}}
& Multi-lingual BERT           & 0.9684 & 0.9658 & 0.9712 & 0.9685 & 0.9145 & 0.8965 & 0.9045 & 0.9004 \\ \cline{2-10}
& XLM-Roberta                  & 0.9524 & 0.9617 & 0.9528 & 0.9572 & 0.9262 & 0.9112 & 0.8904 & 0.9007 \\ \cline{2-10}
& GPT-4                        & 0.8481 & 0.8125 & 0.8847 & 0.8471 & -- & -- & -- & -- \\ \cline{2-10}
& Hybrid Explainability (IIE)  & 0.9315 & 0.9300 & 0.9300 & 0.9300 & 0.8547 & 0.8530 & 0.8525 & 0.8528 \\ \cline{2-10}
& LLM Semi-supervised (LLM-SS) & 0.9220 & 0.9106 & 0.9244 & 0.9174 & 0.7293 & 0.7052 & 0.6951 & 0.7002 \\ \cline{2-10}
& \textbf{Semi-SMDNet}               & \textbf{0.9753} & \textbf{0.9751} & \textbf{0.9748} & \textbf{0.9749} &
                                 \textbf{0.9480} & \textbf{0.9652} & \textbf{0.9480} & \textbf{0.9570} \\ \hline

% ================= Bangla =================
\multirow{6}{*}{\textbf{Bangla}}
& Multi-lingual BERT           & \textbf{0.8284} & 0.6805 & 0.5976 & 0.6364 & 0.7148 & 0.7198 & 0.6145 & 0.6630 \\ \cline{2-10}
& XLM-Roberta                  & 0.8181 & 0.6712 & 0.5407 & 0.6000 & 0.7126 & 0.6412 & 0.5824 & 0.6104 \\ \cline{2-10}
& GPT-4                        & 0.6701 & 0.7200 & 0.6435 & 0.6796 & -- & -- & -- & -- \\ \cline{2-10}
& Hybrid Explainability (IIE)  & 0.7625 & 0.7215 & 0.7602 & 0.7403 & 0.7471 & 0.4910 & 0.5619 & 0.5240 \\ \cline{2-10}
& LLM Semi-supervised (LLM-SS) & 0.7982 & 0.6065 & 0.4921 & 0.5434 & 0.6052 & 0.5822 & 0.5729 & 0.5775 \\ \cline{2-10}
& \textbf{Semi-SMDNet}               & 0.7734 & \textbf{0.7767} & \textbf{0.7733} & \textbf{0.7741} &
                                 \textbf{0.7302} & \textbf{0.7295} & \textbf{0.7299} & \textbf{0.7297} \\ \hline

% ================= English =================
\multirow{6}{*}{\textbf{English}}
& Multi-lingual BERT           & \textbf{0.9148} & \textbf{0.9450} & 0.8100 & 0.8733 & 0.8324 & \textbf{0.8479} & 0.8465 & 0.8472 \\ \cline{2-10}
& XLM-Roberta                  & 0.9072 & 0.8504 & 0.8415 & 0.8459 &
                                 0.8211 & 0.8275 & 0.8355 & 0.8315 \\ \cline{2-10}
& GPT-4                        & 0.8171 & 0.7805 & 0.8780 & 0.8265 & -- & -- & -- & -- \\ \cline{2-10}
& Hybrid Explainability (IIE)  & 0.8572 & 0.8600 & 0.8600 & 0.8600 & 0.6858 & 0.6880 & 0.6880 & 0.6880 \\ \cline{2-10}
& LLM Semi-supervised (LLM-SS) & 0.8631 & 0.8483 & 0.7365 & 0.7885 & 0.6916 & 0.6653 & 0.6430 & 0.6540 \\ \cline{2-10}
& \textbf{Semi-SMDNet}               & 0.8850 & 0.8850 & \textbf{0.8850} & \textbf{0.8850} & \textbf{0.8426} & 0.8415 & \textbf{0.8645} & \textbf{0.8528} \\ \hline

% ================= Spanish =================
\multirow{6}{*}{\textbf{Spanish}}
& Multi-lingual BERT           & 0.9550 & 0.9550 & 0.9570 & 0.9560  &
                                 0.8475 & 0.8478 & 0.8475 & 0.8477  \\ \cline{2-10}
& XLM-Roberta                  & 0.9682 & 0.9369 & \textbf{0.9960} & 0.9657 & 0.9126 & 0.8994 & 0.9025 & 0.9009 \\ \cline{2-10}
& GPT-4                        & 0.7562 & 0.6834 & 0.9604 & 0.7991 & -- & -- & -- & -- \\ \cline{2-10}
& Hybrid Explainability (IIE)  & 0.8767 & 0.8800 & 0.8800 & 0.8800 & 0.8324 & 0.8325 & 0.8325 & 0.8325 \\ \cline{2-10}
& LLM Semi-supervised (LLM-SS) & 0.9475 & 0.9182 & 0.9758 & 0.9458 & 0.7302 & 0.7295 & 0.7299 & 0.7297 \\ \cline{2-10}
& \textbf{Semi-SMDNet}               & \textbf{0.9798} & \textbf{0.9709} & 0.9800 & \textbf{0.9754}& \textbf{0.9254} & \textbf{0.9315} & \textbf{0.9065} & \textbf{0.9188} \\ \hline

\end{tabular}
\label{tab:results}
\end{table*}

\subsection{Performance Analysis}

The semi-supervised framework which is being proposed shows in general and major changes that go in the positive direction for all the four languages, i.e., Arabic, Bangla, English, and Spanish for 100\% and 20\% labelled data, thus pointing out its flexibility to both kinds of settings, i.e., high-resource and low-resource. In contrast to current models that are dependent on large amounts of labelled data, our method is capable of combining unlabelled samples more efficiently by means of pseudo-label refinement, uncertainty weighting, and ensemble learning and therefore, it achieves stronger cross-lingual generalisation.

\subsubsection{Arabic}
In fact, Semi-SMDNet is able to beat the performance of all baseline models across all Arabic datasets within both resource settings. Its F1 score at full data is the highest (0.9749) and, therefore, it outperforms powerful baselines such as M-BERT (0.9685) and XLM-R (0.9572) by a considerable margin. The gap between the best and the second models grows even larger for the 20\% labelled setting, at which point Semi-SMDNet records an F1 of 0.9570 and thus, it is able to outperform the next best model by a big margin. The present findings serve to confirm the strength of the semi-supervised approach envisaged in the paper as a means of dealing with morphologically complex languages such as Arabic.

\subsubsection{Bangla}
Bangla poses the most difficult scenario out of the four languages, which is the main reason why all models have comparatively lower scores. Multi-lingual BERT is the one to deliver the highest accuracy (0.8284) under the 100\% labelled condition; nevertheless, Semi-SMDNet is the one to achieve the best metric balance by recording the highest precision (0.7767), recall (0.7733), and F1 (0.7741) values. At 20\% labelled setting, Semi-SMDNet is the only one to beat the baselines by a large margin with an F1 score of 0.7297, leaving M-BERT (0.6630) and XLM-R (0.6104) far behind. It is worthy of note that other semi-supervised techniques, e.g., LLM-SS and IIE, experience significant drops in their performances while Semi-SMDNet stays steady. This is a clear indication that our method is very effective in situations of low-resource, morphologically rich languages like Bangla.

\subsubsection{English}
All models for English show relatively strong performance, which is consistent with the availability of rich pretraining corpora. M-BERT is the most accurate model (0.9148) under full supervision, while Semi-SMDNet is still a very good performance with a balanced precision, recall, and F1 (0.8850). Semi-SMDNet achieves the highest F1 score (0.8528) in the low-resource setting, thus, it slightly outperforms M-BERT (0.8472) and has better recall performance. The result confirms that the semi-supervised approach they proposed is still efficient in a resource-rich language scenario with limited annotated data.

\subsubsection{Spanish}
Spanish exhibits the highest overall performance across all languages. With 100\% labelled data, Semi-SMDNet achieves the best results across all metrics, attaining an F1 of 0.9754 and improving over strong baselines such as XLM-R (0.9657). In the 20\% labelled setting, it maintains its lead with an F1 score of 0.9188, outperforming M-BERT (0.8477) by a wide margin. These results suggest that Spanish benefits considerably from the model’s cross-lingual representation learning and consistency regularisation, enabling strong generalisation even with reduced supervision.

\medskip

Overall, the language-wise analysis shows that the proposed model not only excels in high-resource scenarios but also significantly mitigates performance degradation in low-resource conditions. The consistent gains across all four languages underscore the importance of ensemble pseudo-labelling, uncertainty weighting, and gradual incorporation of reliable unlabeled samples, making the framework highly suitable for multilingual mental health monitoring where labelled data is limited or expensive to obtain.

\subsection{Ablation Analysis}

To assess the contribution of each major component within the proposed semi-supervised framework, we conducted an ablation study across Arabic, Bangla, English, and Spanish datasets for 100\% labelled data. Four model variants were evaluated: 
removal of data augmentation, removal of Incremental Pseudo-Labeling, 
removal of uncertainty weighting, and 
removal of ensemble learning. 
The performance of these variants is compared against the full model, as reported in Table~\ref{tab:results}, to quantify the effect of each component.

\begin{figure} % Use [H] to force placement
    \centering
\includegraphics[width=0.8\linewidth]{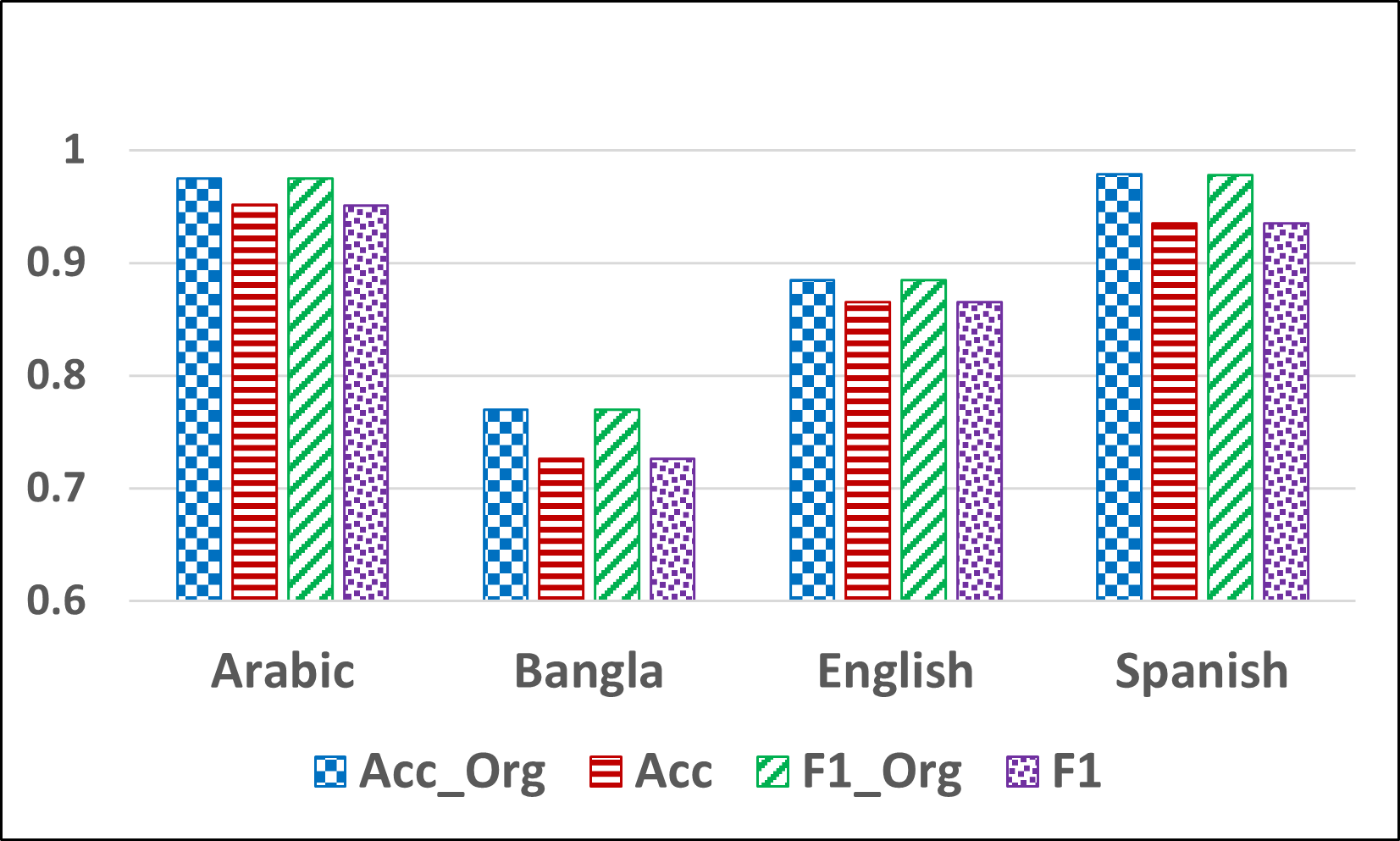}
    \caption{Ablation on data augmentation. Acc\_Org and F1\_Org represent the accuracy and F1-score of the final model. Acc and F1 represent the values after removing the component.}
    \label{fig:abl_aug}
\end{figure}

\paragraph{Effect of Data Augmentation
}
Removing data augmentation led to a moderate performance drop across all languages, with the average F1-score decreasing by approximately 2–4\%. For instance, Arabic declined from 0.974 to 0.951, and Spanish from 0.978 to 0.935. This confirms that augmentation improves the model’s ability to handle noisy and linguistically diverse social media text, particularly for morphologically rich languages such as Arabic and Spanish. However, the Bangla dataset, being low-resource, exhibited relatively minor degradation (0.770 to 0.726), suggesting that augmentation complements but cannot substitute pseudo-labelling in resource-scarce settings. Results are shown in \autoref{tab:abl_aug} and \autoref{fig:abl_aug}.
\begin{table}[h!]
\centering
\caption{ABlation: Effect of Data Augmentation}
\begin{tabular}{|l|l|c|c|c|c|}
\hline
\textbf{Language}  & \textbf{Acc} & \textbf{P} & \textbf{R} & \textbf{F1} \\ \hline
Arabic  & 0.952 & 0.952 & 0.951 & 0.951 \\ \hline
Bangla  & 0.726 & 0.719 & 0.753 & 0.726 \\ \hline
English & 0.865 & 0.867 & 0.869 & 0.865 \\ \hline
Spanish  & 0.935 & 0.935 & 0.937 & 0.935 \\ \hline
\end{tabular}
\label{tab:abl_aug}
\end{table}

\begin{figure}[H] % Use [H] to force placement
    \centering
\includegraphics[width=0.8\linewidth]{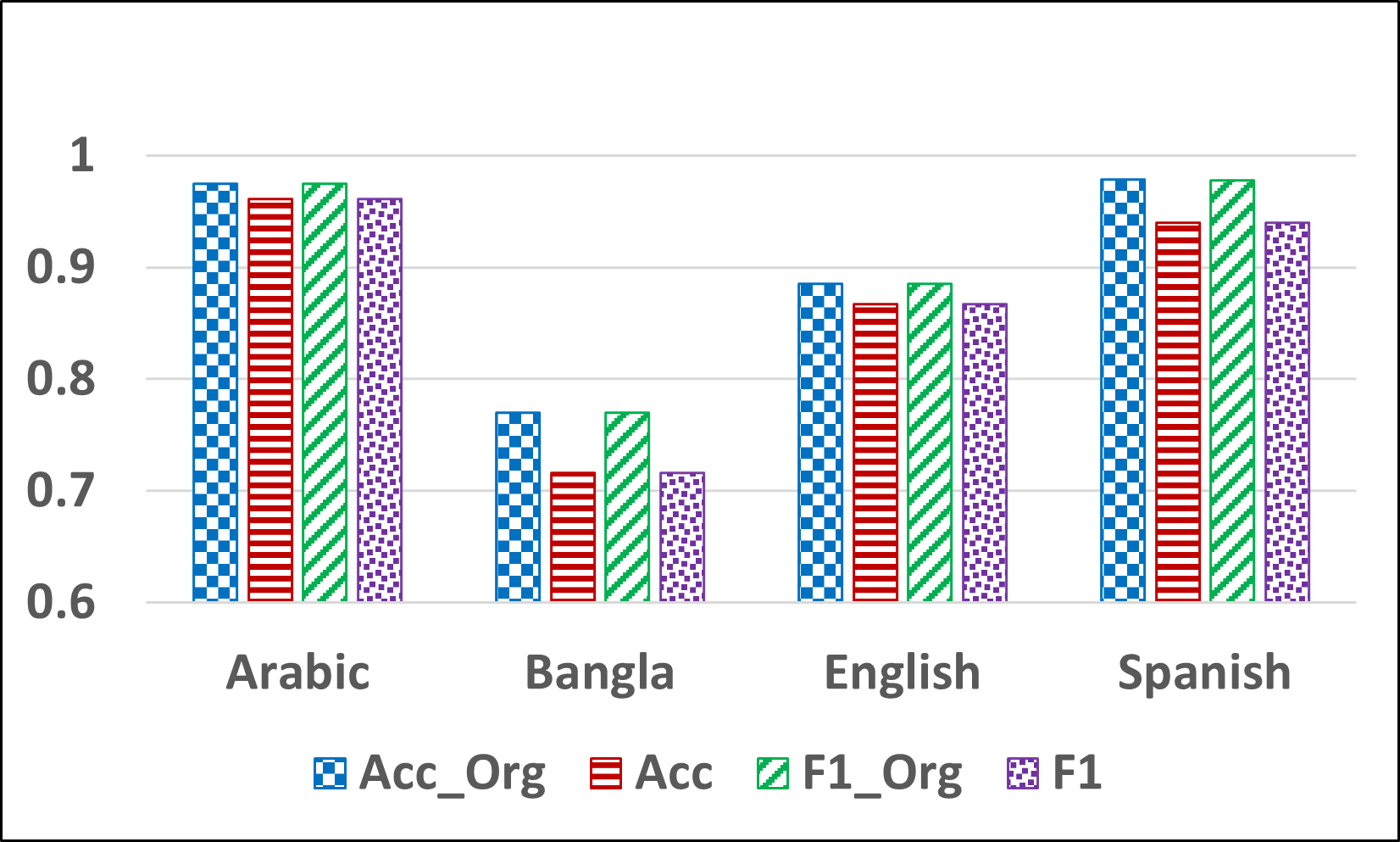}
    \caption{Ablation on IPL. Acc\_Org and F1\_Org represent the accuracy and F1-score of the final model. Acc and F1 represent the values after removing the component.}
    \label{fig:abl_ipl}
\end{figure}

\paragraph{Effect of Incremental Pseudo-Labelling}
When the IPL mechanism was removed, performance decreased for all languages, most notably for Bangla and English, where F1-scores dropped to 0.716 and 0.867, respectively. The IPL strategy plays a crucial role in progressively refining the pseudo-labelled pool by adding high-confidence samples over training iterations. Its absence leads to reduced generalisation and less effective utilisation of unlabeled data. This result highlights IPL as one of the most impactful components for balancing learning stability and label reliability, especially in low-resource and multilingual contexts. Results are shown in \autoref{tab:abl_ipl} and \autoref{fig:abl_ipl}.
\begin{table}[h!]
\centering
\caption{Ablation: Effect of Incremental Pseudo-Labelling}
\begin{tabular}{|l|l|c|c|c|c|}
\hline
\textbf{Language}  & \textbf{Acc} & \textbf{P} & \textbf{R} & \textbf{F1} \\ \hline
Arabic  & 0.961 & 0.961 & 0.962 & 0.961 \\ \hline
Bangla   & 0.716 & 0.712 & 0.727 & 0.716 \\ \hline
English  & 0.867 & 0.865 & 0.883 & 0.867 \\ \hline
Spanish  & 0.940 & 0.940 & 0.942 & 0.940 \\ \hline
\end{tabular}
\label{tab:abl_ipl}
\end{table}

\begin{figure}[H] % Use [H] to force placement
    \centering
\includegraphics[width=0.8\linewidth]{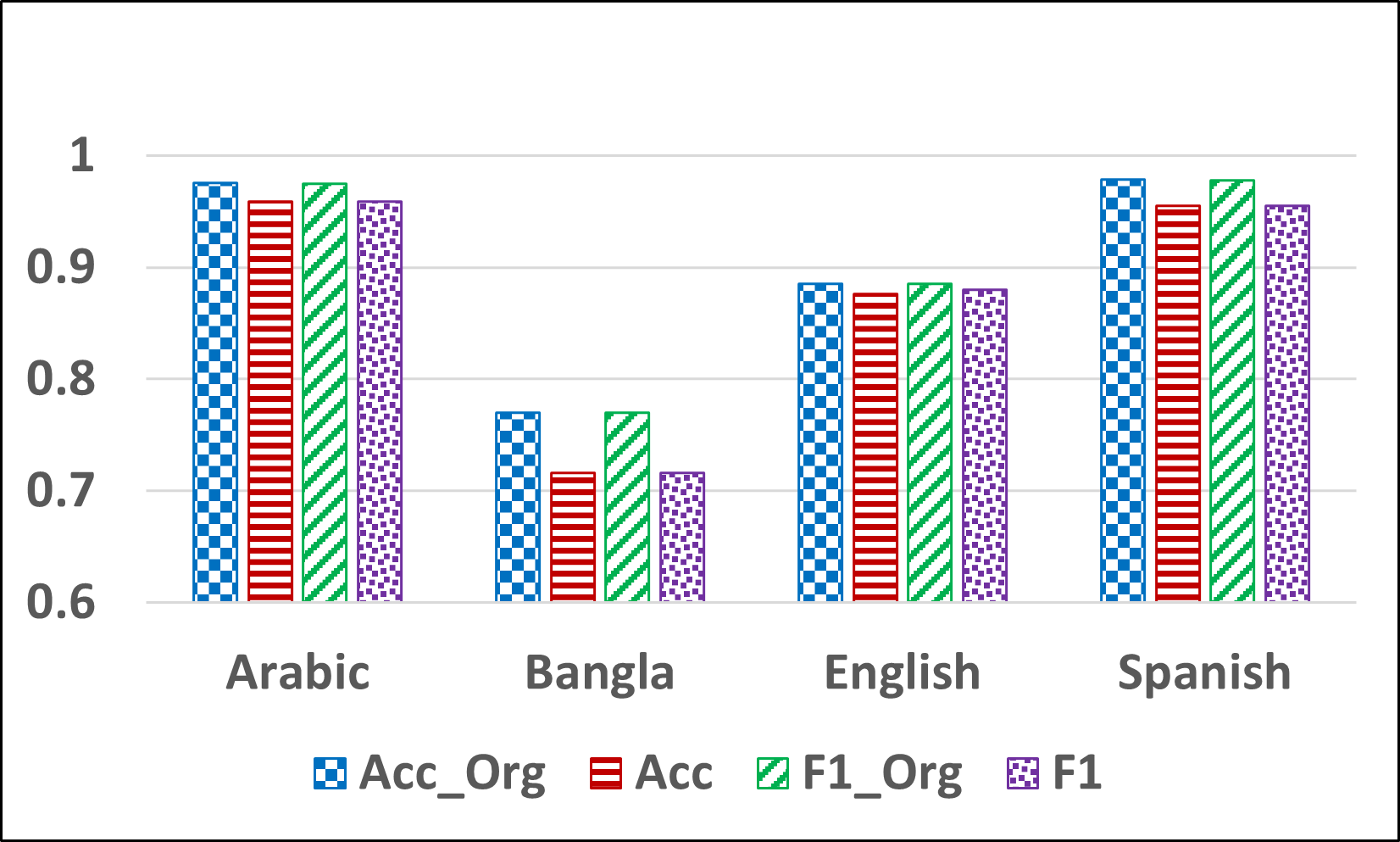}
    \caption{Ablation on uncertainty weighting. Acc\_Org and F1\_Org represent the accuracy and F1-score of the final model. Acc and F1 represent the values after removing the component.}
    \label{fig:abl_unc}
\end{figure}

\paragraph{Effect of Uncertainty Weighting}
The removal of uncertainty weighting slightly reduced the model’s robustness to noisy pseudo-labels. Minor declines were observed in all the languages, emphasising that uncertainty estimation aids in weighting pseudo-labelled samples according to their confidence levels. This mechanism ensures that the student network focuses on more reliable examples, effectively mitigating overfitting to uncertain predictions. Results are shown in \autoref{tab:abl_unc} and \autoref{fig:abl_unc}.
\begin{table}[h!]
\centering
\caption{ABlation: Effect of Uncertainty Weighting}
\begin{tabular}{|l|l|c|c|c|c|}
\hline
\textbf{Language}  & \textbf{Acc} & \textbf{P} & \textbf{R} & \textbf{F1} \\ \hline
Arabic  & 0.959 & 0.959 & 0.959 & 0.959 \\ \hline
Bangla & 0.716 & 0.716 & 0.716 & 0.716 \\ \hline
English & 0.876 & 0.880 & 0.881 & 0.880 \\ \hline
Spanish  & 0.955 & 0.955 & 0.957 & 0.955 \\ \hline
\end{tabular}
\label{tab:abl_unc}
\end{table}

\begin{figure}[H] % Use [H] to force placement
    \centering
\includegraphics[width=0.8\linewidth]{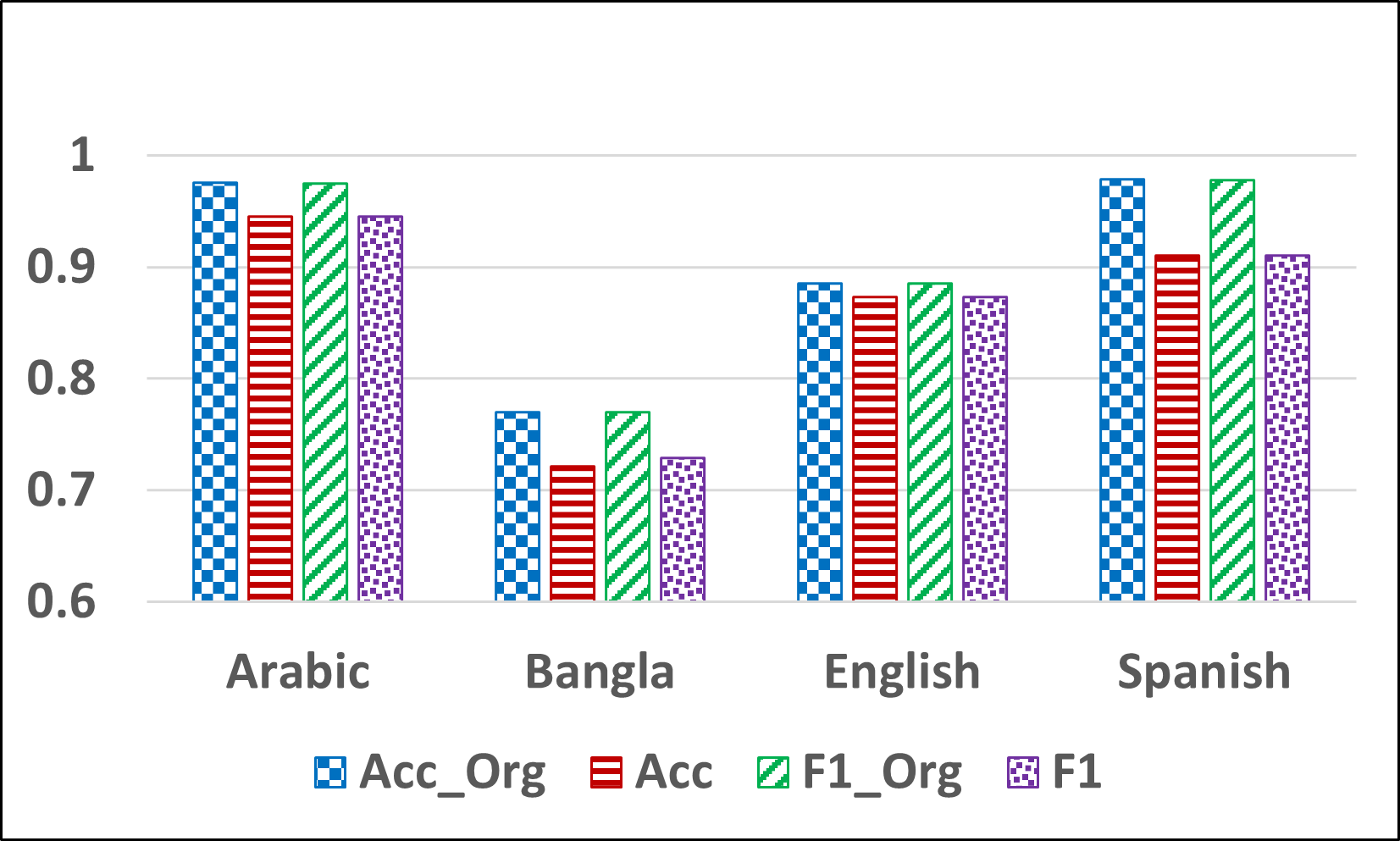}
    \caption{Ablation on ensemble learning. Acc\_Org and F1\_Org represent the accuracy and F1-score of the final model. Acc and F1 represent the values after removing the component.}
    \label{fig:abl_ens}
\end{figure}

\paragraph{Effect of Ensemble Learning}
Excluding the ensemble learning component caused the most significant performance decline across all datasets. For example, Arabic fell from 0.974 to 0.945, and Spanish from 0.955 to 0.910, indicating that ensemble-based aggregation substantially enhances model stability and cross-lingual consistency. Ensemble learning effectively combines multiple pseudo-labelled predictions to reduce bias from any single model, thereby improving decision confidence and mitigating the noise amplification often seen in semi-supervised settings. The results clearly demonstrate that ensemble learning is the most critical component in achieving high generalisation across heterogeneous linguistic distributions. Results are shown in \autoref{tab:abl_ens} and \autoref{fig:abl_ens}.
\begin{table}[h!]
\centering
\caption{Ablation: Effect of Ensemble Learning}
\begin{tabular}{|l|l|c|c|c|c|}
\hline
\textbf{Language}  & \textbf{Acc} & \textbf{P} & \textbf{R} & \textbf{F1} \\ \hline
Arabic   & 0.945 & 0.945 & 0.945 & 0.945 \\ \hline
Bangla   & 0.721 & 0.712 & 0.748 & 0.729 \\ \hline
English  & 0.873 & 0.873 & 0.878 & 0.873 \\ \hline
Spanish  & 0.910 & 0.910 & 0.913 & 0.910 \\ \hline
\end{tabular}
\label{tab:abl_ens}
\end{table}

The ablation analysis reveals that each component contributes uniquely to the model’s overall performance. Ensemble learning and Incremental Pseudo-Labeling (IPL) emerged as the most influential components, followed by data augmentation and uncertainty weighting. Together, these modules enable the framework to achieve a balance between accuracy, robustness, and adaptability, particularly across languages with varying resource availability. The results validate the synergistic design of the proposed semi-supervised architecture and underline the necessity of each component for multilingual depression detection.

\section{Conclusion}
\label{sec:sa_cnfw}

In this paper, we propose, Semi-SMDNet, a semi-supervised framework for multilingual depression detection on social media. The model addresses key challenges such as multilingualism, code-mixed text, and limited labeled data by combining transformer-based text embeddings with social context features like user interactions and content polarity in a unified architecture. Our approach leverages cross-lingually trained embeddings and pseudo-labeled data to enhance performance, especially for low-resource languages. The integration of textual and behavioral cues enables the model to capture nuanced depressive signals across diverse linguistic settings. Extensive experiments on large multilingual datasets demonstrate that the proposed method outperforms existing baselines. The results confirm the value of combining text and context features, as well as ensemble learning, for scalable and inclusive depression detection.

\balance

\bibliographystyle{IEEEtran}
\bibliography{references}

% \vfill

\end{document}